\documentclass[runningheads]{llncs}
\usepackage{graphicx}
\usepackage{comment}
\usepackage{amsmath,amssymb} 
\usepackage{color}
\usepackage{bbm}
\usepackage{hyperref}

\DeclareMathOperator{\GPR}{GPR}

\newcommand{\vvec}[1]{\mathbf{#1}}
\newcommand{\pheadB}[1] {\vspace{1mm}\noindent\textbf{#1}}

\makeatletter
\newcommand{\printfnsymbol}[1]{%
  \textsuperscript{\@fnsymbol{#1}}%
}
\makeatother

\begin{document}
	\pagestyle{headings}
	\mainmatter
	\def\ECCVSubNumber{1380}  
	
	\title{Graph-PCNN: Two Stage Human Pose Estimation with Graph Pose Refinement } 
	
	\titlerunning{Graph-PCNN}
	%
	\author{Jian Wang\printfnsymbol{1} \and
    Xiang Long\thanks{Both authors contributed equally to this work.} \and
	Yuan Gao \and
	Errui Ding \and
	Shilei Wen
	}
	\authorrunning{J. Wang, X. Long et al.}
	%
	\institute{Department of Computer Vision Technology(VIS), Baidu Inc.\\
    \email{ \tt\small \{wangjian33,longxiang,gaoyuan18,dingerrui,wenshilei\}@baidu.com}
	}
	\maketitle
	
	\begin{abstract}
		Recently, most of the state-of-the-art human pose estimation methods are based on heatmap regression. The final coordinates of keypoints are obtained by decoding heatmap directly. In this paper, we aim to find a better approach to get more accurate localization results. We mainly put forward two suggestions for improvement: 1) different features and methods should be applied for rough and accurate localization, 2) relationship between keypoints should be considered.
		Specifically, we propose a two-stage graph-based and model-agnostic framework, called Graph-PCNN, with
		a localization subnet and a graph pose refinement module added onto the original heatmap regression network. In the first stage, heatmap regression network is applied to obtain a rough localization result, and a set of proposal keypoints, called guided points, are sampled. In the second stage, for each guided point, different visual feature is extracted by the localization subnet. The relationship between guided points is explored by the graph pose refinement module to get more accurate localization results.
		Experiments show that Graph-PCNN can be used in various backbones to boost the performance by a large margin.  
		Without bells and whistles, our best model can achieve a new state-of-the-art 76.8\% AP on COCO \texttt{test-dev} split.

		\keywords{Human Pose Estimation, Keypoint Localization, Two Stage, Graph Pose Refinement}
	\end{abstract}

	\section{Introduction}
	
	Human pose estimation\cite{andriluka20142d} is a fundamental yet challenging computer vision problem, that aims to localize keypoints (human body joints or parts). It is the basis of other related tasks and various downstream vision applications, including video pose estimation\cite{xiaohan2015joint}, tracking\cite{cho2013adaptive,xiao2018simple} and human action recognition \cite{liang2014expressive,wang2013approach,yan2018spatial}.
	This paper is interested in 2D pose estimation to detect the spatial location (i.e. 2D coordinate) of keypoints for persons in a top-down manner.
	Keypoint localization is a very challenging task, even for humans. It is really difficult to locate the keypoint coordinates precisely, since the variation of clothing, the occlusion between the limbs, the deformation of human joints under different poses and the complex unconstrained background, will affect the keypoint recognition and localization \cite{zhang2019distribution}.
	
	\begin{figure} [t]
		\centering
		\includegraphics[height=0.5\textwidth]{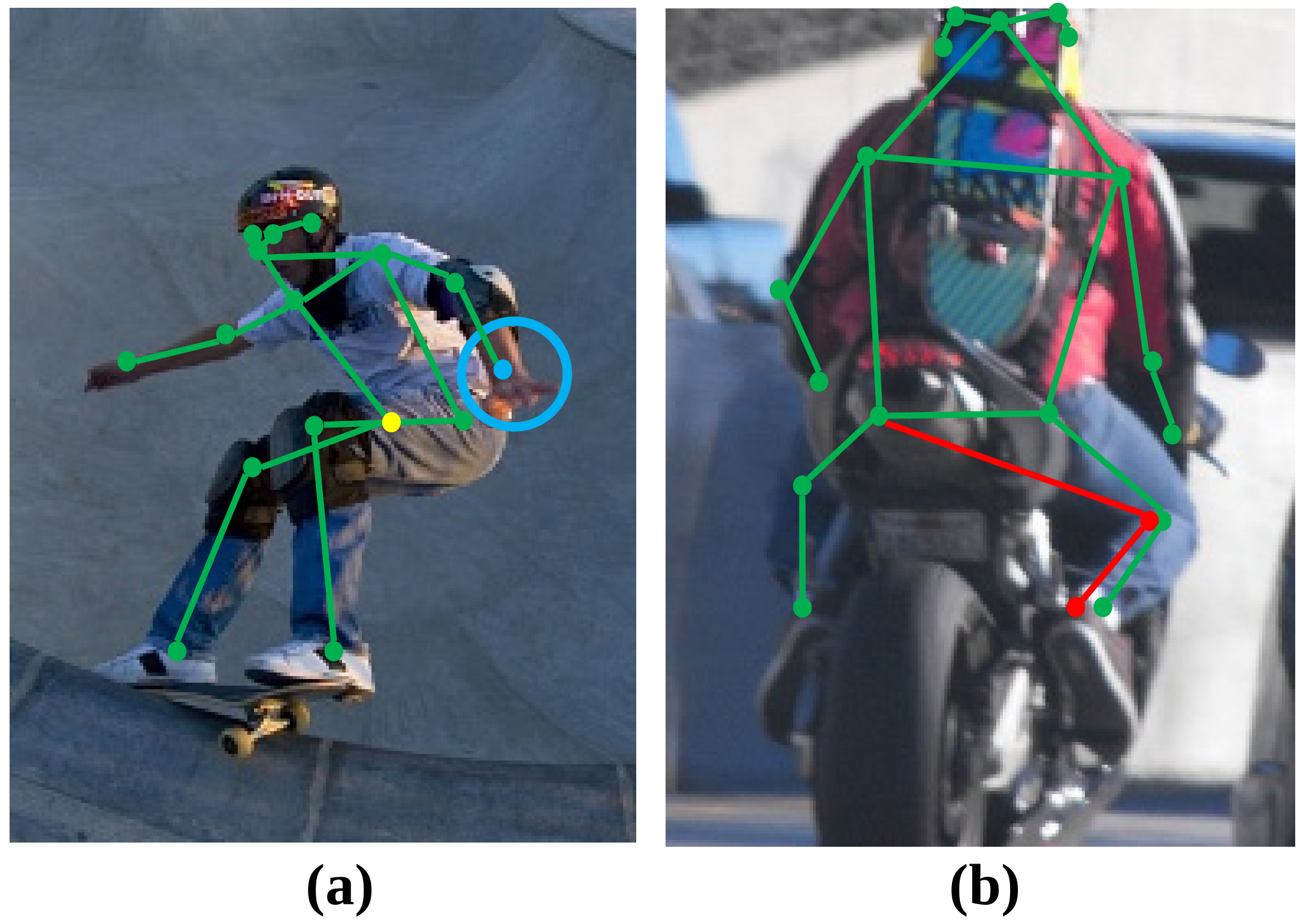}
		\caption{\textbf{Example of 2D pose estimation}. The green points and lines indicate keypoints and their connections that are correctly predicted, while the red ones indicate incorrect predictions.
			We have observed two important characteristics of keypoint localization: 1) different features and processes are preferred for rough and accurate localization, 2) relationship between keypoints should be considered.}
		\label{fig:intro}
	\end{figure}
	
	Most existing state-of-the-art methods use CNNs to get the heatmap of each keypoint	
	\cite{chen2014articulated,gkioxari2016chained,wei2016convolutional,chu2016crf,chu2017multi,yang2016end,chu2016structured,sun2017compositional,yang2017pyramid,xiao2018simple,ning2017tmm,tang2018deeply,chen2018cascaded,ke2018multi,liu2018cascaded,su2019multi,sun2019deep}. Then the heatmap will be directly decoded to keypoint coordinates. 
	However, these approaches do not take into account two important characteristics of human pose estimation: 1) different features and processes are preferred for rough and accurate localization, 2) relationship between keypoints should be considered.
	
	First, humans perform keypoint localization in a two-step manner\cite{fieraru2018learning}, \cite{bulat2016human}, \cite{belagiannis2015robust}, \cite{moon2019posefix}. For example, for the blue point in Fig.~\ref{fig:intro}(a), we will first perform a rough localization based on the context information, including fingers and arms shown in the blue circle, to determine whether there is a wrist keypoint in a nearby area. This step can be treated as a proposal process. After rough localization, we will further observe the detail structure of the wrist itself to determine the accurate location of wrist keypoint, which can be seen as a refinement process.
	
	We get inspiration from object detection. In object detection methods, proposal and refinement are performed based on two different feature map achieved by two separate subnets. We suggest that the proposal and refinement processes in keypoint localization should also be based on different feature maps. Therefore, we apply two different subnets to get feature maps for proposal and refinement respectively. Besides, two-stage method is very common in object detection, and can achieve excellent results in terms of both effectiveness and performance. A natural idea is that we apply the design of the two-stage to keypoint localization task, let the first stage focus on the proposal process, improving the recall of keypoint, and let the second stage focus on the refinement process, improving the localization accuracy.  Therefore, we introduce the concept of Guided Point. First, we select the guided points based on the heatmap as rough proposals in the first stage. Then in the second stage, based on the corresponding features of selected guided points, we perform coordinate refinement for accurate keypoint regression.
	
	Secondly, in the case of complicated clothing and occlusion, the relationship between the keypoints is very important to judge its location. For example, the yellow keypoint in Fig.~\ref{fig:intro}(a), due to the occlusion, we cannot see its location directly. We can only infer it from the location of other related keypoints. In addition, due to the structural limitations of the human body, there exist obvious mutual constraints between keypoints. In the refinement process, considering the relationship between keypoints may help to avoid and correct the misprediction. For example, in Fig.~\ref{fig:intro}(b), the keypoints in red are the wrong predictions of the left leg. By considering the connection between them and other keypoints, we can find out and correct these errors more easily.
	
	However, in the traditional heatmap based method, we cannot know the location of keypoints before decoding the heatmap to coordinates. 
	This makes it difficult for us to build a pose graph that connects keypoint features at different locations. 
	After introducing guided points, we can know the rough locations of keypoints, such that we can build a pose graph between keypoints easily. 
	Therefore, we propose a graph pose refinement (GPR) module, which is an extension of graph convolutional network, to improve the accuracy of keypoint localization.

	The main contributions of this paper include:
	\begin{itemize}
		\item This paper proposes a model-agnostic two-stage keypoint localization framework, Graph-PCNN, which can be used in any heatmap based keypoint localization method to bring significant improvement.
		
		\item A graph pose refinement module is proposed to consider the relationship between keypoints at different locations, and further improve the localization accuracy.
		
		\item Our method set a new stage-of-the-art on COCO \texttt{test-dev} split.
	\end{itemize}

	\section{Related work}
	
	The classical approach of human pose estimation is using the pictorial structures framework with a pre-defined pose or part templates not depending on image data, which limit the expressiveness of the model \cite{yang2012articulated,pishchulin2013poselet}. 
	
	Convolution Neural Networks (CNNs) have dramatically changed the direction of pose estimation methods. Since the introduction of "DeepPose" \cite{toshev2014deeppose} by Toshev et al., most recent pose estimation systems have generally adopted CNNs as their backbone.  
	There are mainly two kinds of methods to get the locations of keypoints: directly regress coordinates and estimate the heatmaps of the keypoints first, and then decode to coordinates.
	
	\pheadB{Coordinate based Methods} 
	Only a few methods regress coordinates of keypoints directly. DeepPose \cite{toshev2014deeppose} formulate pose estimation as a CNN-based regression problem directly towards body joints in a holistic fashion.  Fan et al., \cite{fan2015combining} propose to integrate both the body part appearance and the holistic view of each local part for more accurate regression. A few other methods \cite{carreira2015human,sun2018integral} further improve performance, but there is still a gap between with heatmap based methods.
	
	\pheadB{Heatmap based Methods} 
	The heatmap representation is first introduce by Tompson et al. \cite{tompson2014joint}, 
	and then quickly becomes the most popular solution in state-of-the-art methods.
	A lot of research works improve the network  architectures to improve the effectiveness of heatmap regression~\cite{chen2014articulated,gkioxari2016chained,belagiannis2016recurrent,lifshitz2016human,newell2016stacked,wei2016convolutional,chu2016crf,chu2017multi,yang2016end,chu2016structured,sun2017compositional,yang2017pyramid} \cite{xiao2018simple,ning2017tmm,tang2018deeply,chen2018cascaded,ke2018multi,liu2018cascaded,su2019multi,sun2019deep}. For example, Hourglass \cite{newell2016stacked} and its follow-ups \cite{yang2017pyramid,chen2017adversarial,chu2017multi} consist of blocks of several pooling and upsampling layers, which looks like an hourglass, to capture information at every scale. 
	SimpleBaseline \cite{xiao2018simple} adds several deconvolutional layers to enlarge the resolution of output feature maps, which is quite simple but performs better.
	The HRNet \cite{sun2019deep} model has outperformed all existing methods on public dataset by maintaining a high-resolution representation through the whole process.
	
	\pheadB{Hybrid Methods}
	Some works speculate that heatmap will introduce a statistical error and try to combine heatmap estimation with coordinate offset regression for better localization accuracy \cite{papandreou2017towards,huang2019devil}. But in these methods, heatmap estimation and coordinate regression are performed at the same time on the same feature map, without refinement process to gradually improve accuracy.
	
	\pheadB{Refinement Methods} 
	Many works focus on coordinate refinement to improve the accuracy of keypoints localization\cite{carreira2015human,bulat2016human,fieraru2018learning,moon2019posefix}. Instead of predicting absolute joint locations, Carreira et al. refine pose estimation by predicting error feedback at each iteration\cite{carreira2015human}, Bulat et al. design a cascaded architecture for mining part relationships and spatial context\cite{bulat2016human}. Some other works use a human pose refinement network to exploit dependencies between input and output spaces \cite{fieraru2018learning,moon2019posefix}.
	However, they can not effectively combine heatmap estimation and coordinate regression, and the relationship between different keypoints is not considered during refinement. 
	Our method will introduce the relationship between keypoints for more effective refinement.
	Zhang et al. \cite{zhang2019human} builds a pose graph directly on heatmaps and uses Graph Neural Network for refinement. However, it essentially only considers the relationship between heatmap weights at the same location, while the visual information of keypoints is completely ignored. In our framework, pose graph is built on the visual features at the position of corresponding keypoints, which is more conducive to subsequent refinement.

	\section{Two stage pose estimation framework}
	
	\begin{figure} [t]
		\centering
		\includegraphics[width=1.0\textwidth]{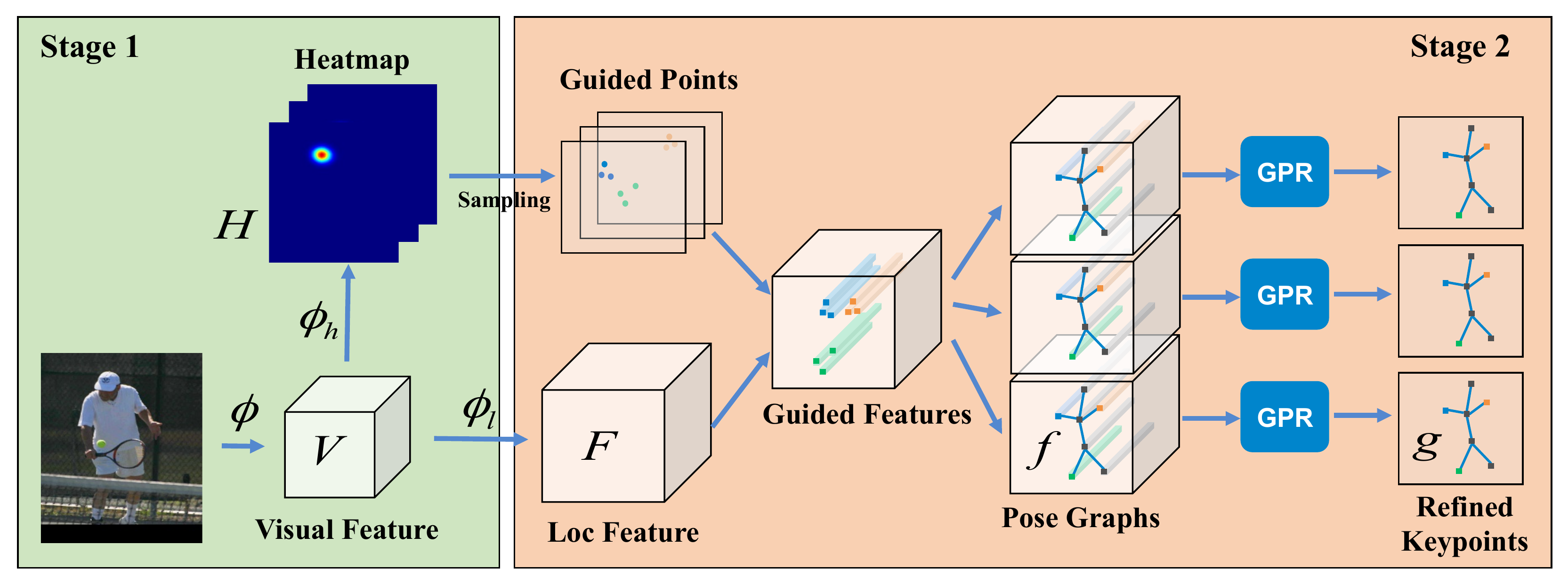}
		\caption{\textbf{Overall architecture of two stage pose estimation framework}. In the first stage, heatmap regressor is applied to obtain a rough localization heatmap, and a set of guided points are sampled. In the second stage, guided points with corresponding localization features are constructed as pose graphs and then feed into a graph pose refinement (GPR) module to get refined results.}
		\label{fig:method_overview}
	\end{figure}
	
	In the top-down manner pose estimation methods, single person pose estimator aims to locate $K$ keypoints $\vvec{P} = \{\vvec{p}_1, \vvec{p}_2, ..., \vvec{p}_k\}$ from an image $\vvec{I}$ of size $W \times H \times 3$, where $\vvec{p}_k$ is a 2D-coordinates. Heatmap based methods transform this problem to estimating $K$ heatmaps $\{ \vvec{H}_1, \vvec{H}_2, ... , \vvec{H}_k \} $ of size $W' \times H' \times K$, where each heatmap $\vvec{H}_k$ will be decoded to the corresponding coordinates $\vvec{p}_k$ during the test phase.  
	
	Our method simply follows the popular methods to generate the heamap in the first stage. A common pipeline is first to use a deep convolutional network $\phi$ to extract visual features $\vvec{V}$ from image $\vvec{I}$,
	\begin{equation}
	\vvec{V} = \phi(\vvec{I}).
	\end{equation}
	A heapmap regressor $\phi_h$, typically ended with a $1 \times 1$ convolutional layer, is applied to estimating the heatmaps,
	\begin{equation}
	\{ \vvec{H}_1, \vvec{H}_2, ... , \vvec{H}_k \} = \phi_h(\vvec{V}).
	\end{equation}
	
	The refinement network is added after the heatmap regression, without any changes to the existing network architecture in the first stage. Therefore, our method can be applied to any heatmap based models easily. The overall architecture of our method is shown in Fig.~\ref{fig:method_overview}.
	At first, we apply a localization subnet $\phi_l$ to transform the visual feature to the same spacial scale as heatmaps,
	\begin{equation}
	\vvec{F} = \phi_l(\vvec{V}),
	\end{equation}
	where the size of $\vvec{F}$ is $W' \times H' \times C$.
	During training, $N$ guided points $\{ \vvec{s}^1_k, \vvec{s}^2_k, ..., \vvec{s}^N_k \}$ are sampled for each heatmap $\vvec{H}_k$, while the best guided points $\vvec{s}^*_k$ is selected for heapmap $\vvec{H}_k$ during testing. For sake of simplification, we omit the superscript in the following formula. For any guided point $\vvec{s}_k$, guided feature $\vvec{f}_k=\vvec{F} [\vvec{s}_k]$ at the corresponding location and its confidence score $h_k=\vvec{H_k} [\vvec{s}_k]$ can be extracted.
	
	Subsequently,  we can build $N$ pose graph for $N \times K$ guided features, and introduce a graph pose refinement (GPR) module to refine the visual features by considering the relationship between keypoints.  
	\begin{equation}
	\{\vvec{g}_1, \vvec{g}_2, ...,  \vvec{g}_K \} = \GPR(\{ \vvec{f}_1, \vvec{f}_2, ...,  \vvec{f}_K \}, \{ h_1, h_2, ..., h_K \} ).
	\end{equation}	
	Finally, the refined classification result $\vvec{c}_k$ and offset regression result $\vvec{r}_k$ are achieved based on the refined feature $\vvec{g}_k$. The refined coordinate of keypoint is
	\begin{equation}
	\vvec{p}_k = \vvec{s}_k + \vvec{r}_k .
	\end{equation}
	
	
	In the following, we first describe the guided point sampling strategy in section \ref{sec:sampling}. Second, we show the detail structure of graph pose refinement module in section \ref{sec:gcn}. Third, we introduce the loss used for training in section \ref{sec:loss}. Finally, we show how to integrate our framework to existing backbones and elaborate the details of training and testing in section \ref{sec:detail}.

	\subsection{Guided Point Sampling} \label{sec:sampling}
	Locating human joint based on the peak of heatmap is frequently-used in modern human pose estimators, and they modeled the target of heatmap by generating gaussian distribution around the ground truth. But due to the complex image context and human action, the joint heat may not be satisfy gaussian distribution strictly which, together with quantisation affect of image resolution downsampling, leads to an insufficient precision of this localization method. However, the peak of heatmap is always close to the true location of joint, which make it adequate to regress the true location. 
	
	\begin{figure} [t]
		\centering
		\includegraphics[width=1.0\textwidth]{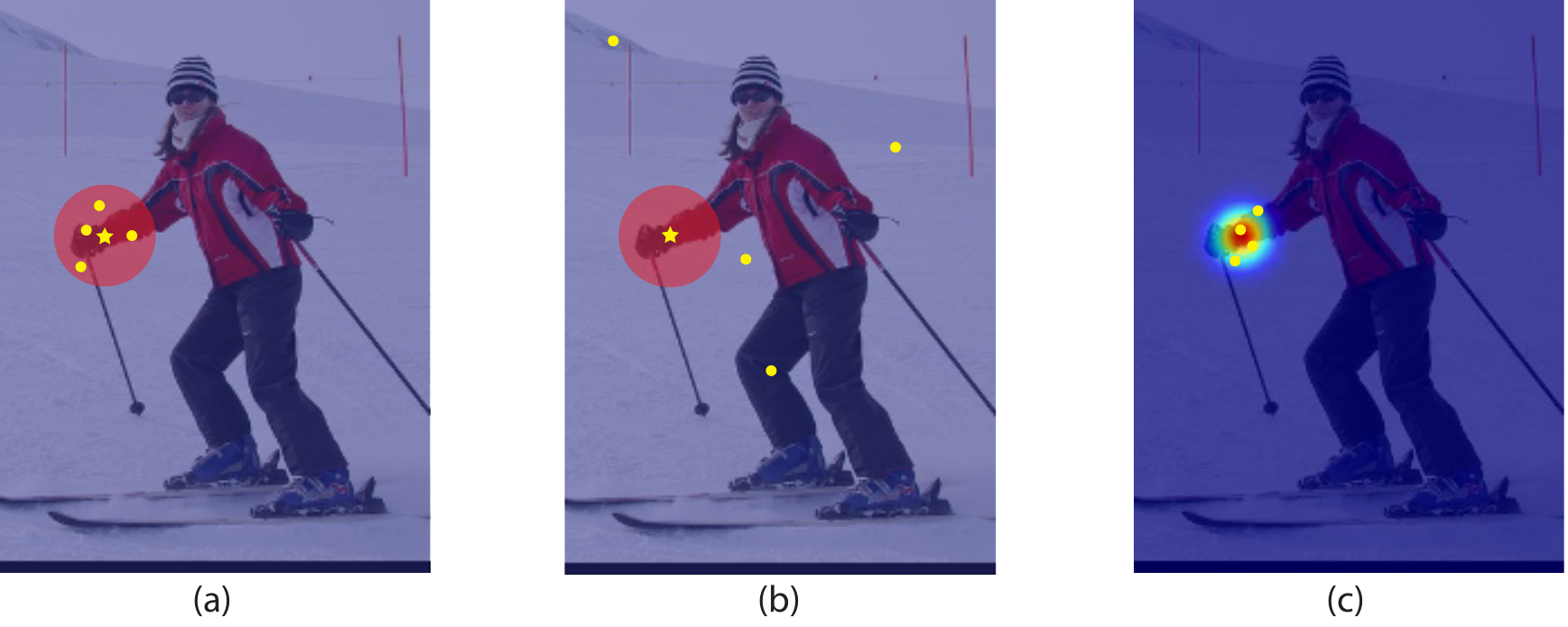}
		\caption{\textbf{Illustration of sampling region}. Taking right wrist as example, (a), (b), (c) show the three kinds of guided points respectively, points which are close to the ground truth keypoint, points which are far away from the ground truth keypoint and points which have high heat response, where the yellow circle points indicate sampled guided points and the yellow star points indicate ground truth of right wrist.}
		\label{fig:proposalsampling}
	\end{figure}
	
	To achieve the goal of obtaining refined coordinate based on the peak of heatmap, we sample several guided points and train coordinate refinement in stage2. Concretely, we equally sample three kinds of guided points for training: (a) points which are close to the ground truth keypoint, (b) points which are far away from the ground truth keypoint, and (c) points which have high heat response. And the $k$th ground truth keypoint is denoted as $\vvec{t}_k$. As exhibited in Fig.~\ref{fig:proposalsampling}, (a) and (b) are randomly sampled within the red region and blue region, respectively, and the red region which centered at ground truth has a radius of $3\sigma$, where $\sigma$ is same with the standard deviation for generating gaussian heatmap target. (c) is randomly sampled from the top $N$ highest response points at the heatmap.
	
	Due to the different characterization of different keypoints, we sample guided points for each keypoint individually, and the total amount of the three kinds of guided points for each keypoint is set equally to $N$.
	
	After the $N$ guided points $\{ \vvec{s}^1_k, \vvec{s}^2_k, ..., \vvec{s}^N_k \}$ are sampled, we divide them into two sets, positive set and negative set, denoted as
	
	\begin{equation}
	\begin{split}
	& \mathcal{S}_k^+ = \{ \vvec{s}_k \ | \  ||\vvec{s}_k-\vvec{t}_k|| < 3\sigma \} \\
	& \mathcal{S}_k^- = \{ \vvec{s}_k \ | \  ||\vvec{s}_k-\vvec{t}_k|| \geq 3\sigma \}
	\end{split}
	\end{equation}
	and $N_k^+=|\mathcal{S}_k^+|$, $N_k^-=|\mathcal{S}_k^-|$. Then all of the corresponding guided feature extracted from $\vvec{F}$ by means of bilinear interpolation are feeded into stage2 for refinement while only the guided points from positive set contributed to the coordinate regression.
	
	According to the above label assignment manner, (a) and (b) are definite positive and negative samples, and the influence of proportion between them will be explored in Section~\ref{sec:ablation}. While (c) is almost negative samples during the beginning stage of training and turns to positive samples as the training schedule goes on. We suppose that (c) can not only accelerate the feature learning at the beginning of training, because (c) are hard negative samples for classification at this stage, but also contribute to the learning of regression when the classification status of feature is relatively stable, as (c) are almost positive samples at this period. Further more, (c) is not necessarily positive when the model converges roughly because of some prediction error caused by hard situation. In this circumstances, (c) can also be regarded as hard negative samples for helping the model to be trained better.
	
	\subsection{Graph Pose Refinement}  \label{sec:gcn}
	
	In most of previous works, many fields have been well studied for human pose estimation, such as network structure, data preprocessing and postprocessing, post refinement, etc. However, in these works, the localization of human keypoints are conducted independently for each keypoint while the relationship between different keypoints is ignored all along. Intuitively, the human keypoints construct a salient graph structure base on the pattern of human body, and they have clear adjacent relation with each other. So we consider that the localization of keypoints can be infered better with the help of the information hinted by this relationship. For instance, in our framework, if we know that a guided point is left elbow, then the positive guided points of left wrist should tend to have higher response on left wrist a priori, as left wrist is adjacent to left elbow. So that more supervision can be imposed upon the feature of these keypoints than treating them independently.

	\begin{figure} [t]
		\centering
		\includegraphics[width=1.0\textwidth]{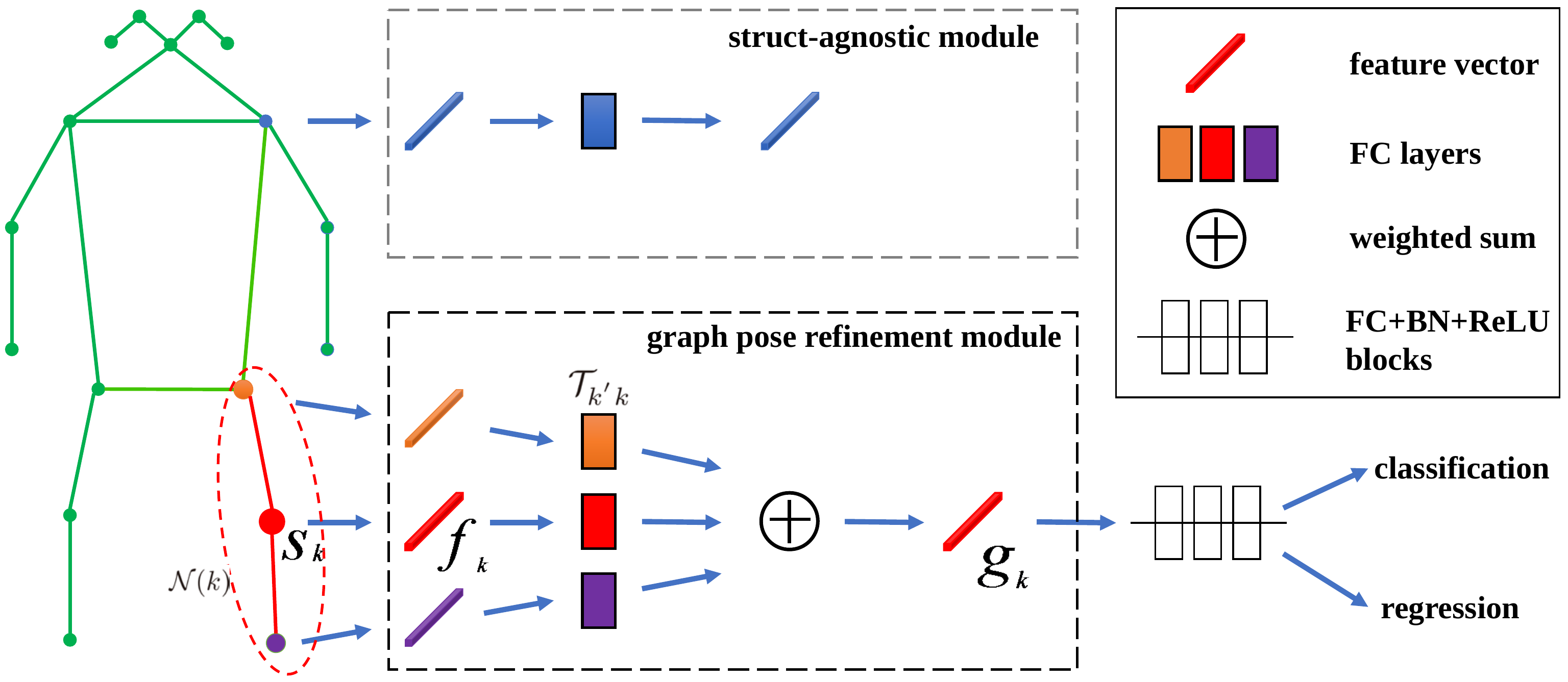}
		\caption{\textbf{The structure of graph pose refinement module}. The relationship between keypoints is taken into account in contrast to the struct-agnostic module.}
		\label{fig:gcnhead}
	\end{figure}

	To take advantage of the information implicit in the graph structure mentioned above, we propose a graph pose refinement module to model it, and then refine the feature of these keypoints. As shown in Fig.~\ref{fig:gcnhead}, we build a graph and conduct graph convolution for each keypoint. The output embedding feature can be computed by 
	
	\begin{equation}
	\label{fml:computegcn}
	\begin{split}
	& \vvec{g}_k = \frac{1}{Z_k}\sum_{\vvec{s}_{k^{'}} \in \mathcal{N}(k)}     \omega_{k^{'}} \mathcal{T}_{k^{'}k}(\vvec{f}_{k^{'}}) \\
	& \omega_{k^{'}} = \left\{\begin{aligned}
	& h_{k^{'}} \mathbbm{1}(R_{k^{'}}), & k^{'} \ne k \\
	& 1, & k^{'} = k
	\end{aligned}\right. 
	\end{split}
	\end{equation}
	where $\mathcal{N}(k)$ represents for a point set containing the guided point $\vvec{s}_k$ and its neighbours, $\mathcal{T}_{k^{'}k}$ for the linear transformation from guided point $\vvec{s}_{k^{'}}$ to $\vvec{s}_k$, and $\mathbbm{1}$ for the indicator function. $Z_k = \sum_{\vvec{s}_{k^{'}} \in \mathcal{N}(k)} \omega_{k^{'}}$ is used for normalization. $R_{k^{'}}$ is a boolean type parameter encoding the reliability of a guided point which works for filtering out points of low quality, and its definition will be explained in detail in section~\ref{sec:detail}.
	
	Specially, as defined in (\ref{fml:computegcn}), this graph convolution is an extension of traditional graph convolution, it is designed by considering the characteristic of pose estimation problem. Firstly, we add a weight for each message generated from $\vvec{s}_k^{'}$ to $\vvec{s}_k$, which can control the contribution of each message according to the intensity and reliability of $\vvec{s}_k^{'}$. With the constraint of these weights, the graph convolution can be trained more stably. Further more, we set $\omega_{k^{'}}=1$ when $k^{'} = k$. And this can make the graph convolution degrading to a traditional linear transformation for $\vvec{s}_k$ when $\mathbbm{1}(R_{k^{'}})=0$ for all $\vvec{s}_{k^{'}} \in \mathcal{N}(k)$ where $k^{'} \ne k$, without being affected by the intensity and reliability of $\vvec{s}_k$ itself.
	
	\subsection{Loss Function}  \label{sec:loss}
	
	After the refinement module above, the embedded feature is sent to a module containing several fully connected layers and batch norm layers, as illustrated in Fig.~\ref{fig:gcnhead}. Finally two predictions are outputed, denoted as $\vvec{c}_k$ and $\vvec{r}_k$, for classification and regression respectively. Giving ground truth keypoint location $\vvec{t}_k$, the losses for these two branches are defined as
	
	\begin{equation}
	\begin{split}
	L_k^{cls} &= \frac{1}{2}  \left[ \frac{1}{N_k^+}\sum_{\vvec{s}_k^i \in \mathcal{S}_k^+} \alpha_k^i\mathcal{L}_{cls}(\vvec{c}_k^i, 1) + \frac{1}{N_k^-}\sum_{\vvec{s}_k^i \in \mathcal{S}_k^-}\mathcal{L}_{cls}(\vvec{c}_k^i, 0) \right] \\
	& \hspace{2.3cm} \alpha_k^i = exp(-\frac{(\vvec{s}_k^i-\vvec{t}_k)^2}{2\sigma^2})
	\end{split}
	\end{equation}
	and 
	\begin{equation}
	L_k^{reg} = \frac{1}{N_k^+}\sum_{\vvec{s}_k^i \in \mathcal{S}_k^+}\mathcal{L}_{reg}(\vvec{r}_k^i, \vvec{t}_k-\vvec{s}_k^i),
	\end{equation}
	where $\mathcal{L}^{cls}$ and $\mathcal{L}^{reg}$ are softmax cross-entropy loss and L1 loss.
	The total loss of the stage2, can be expressed as
	
	\begin{equation}
	L^{s_2} = \frac{\sum_k \gamma_k(L_k^{cls} + \lambda L_k^{reg})}{\sum_k \gamma_k},
	\end{equation}
	where $\gamma_k$ is the target weight of keypoint $k$. And $\lambda$ is a loss weight which is set to 16 constantly. And the total loss of Graph-PCNN is
	
	\begin{equation}
	L = L^{s_1} + L^{s_2},
	\end{equation}
	where $L^{s_1}$ is the traditional heatmap regression loss for stage1.
	
	\subsection{Network Architecture}  \label{sec:detail}
	
	\subsubsection{Network Architecture}
	
	In previous works such as \cite{papandreou2017towards}, \cite{huang2019devil}, there is also a coordinate refinement after heatmap decoding, and their coordinate refinement branch share the same feature map with heatmap prediction branch. However, the rough and accurate localization always need different embedding feature, further more, it is hard to conduct particular feature refinement for either of these two branches. In order to alleviate the above problems, we copy the last stage of the backbone network to produce two different feature maps with the same size followed by heatmap regression convolution and graph pose refinement module respectively. By means of this modification, the network can learn more particular feature for two different branches, and easily conduct guided points sampling for further feature refinement.
	
	\subsubsection{Training and Testing}
	
	For the proposed two stage pose estimation framework, several operations are specific in the training and testing phase.
	
	Firstly, in order to make the stage2 be trained sufficiently, we sample multiple guided points for each keypoint following the strategy described in Section~\ref{sec:sampling} during training, and the amount of guided points $N$ is various according to the input size. While during testing, only one guided point is generated by decoding the predicted heatmap, and the output score of it is gathered as the corresponding heat response score from stage1. Following most of previous works\cite{xiao2018simple}, \cite{sun2019deep}, a quarter offset in the direction from the highest response to the second highest response is added to the position of heatmap peak for higher precision, when decoding this guided point from heatmap.
	
	Secondly, the definition of guided point reliability metric $R_{k^{'}}$ is different for training and testing, which is represented as
	
	\begin{equation}
	R_{k^{'}} = \left\{\begin{aligned}
	& ||\vvec{s}_{k^{'}}-\vvec{t}_{k^{'}}||<\delta & in\ training\ phase \\
	& h_{k^{'}} > \xi & in\ testing\ phase 
	\end{aligned}\right.
	\end{equation}
	At the training phase, the ground truth is available for measuring this reliability, and the guided points which are close to their corresponding ground truth can be regarded reliable. $\delta$ is a distance threshold controling the close degree which equals to $2\sigma$. While at the testing phase the ground truth is unknown, so for insurance, the guided points which heat responses are high enough are qualified to pass message to their neighbour points. And $\xi$ is a threshold for gating the heat response, which is set to 0.85 constantly.
	
	Finally, during training, we shuffle the guided points of one keypoint after the guided point sampling in order to create more various situation of graph combination, which can make the graph pose refinement module more generalized.

	\section{Experiments}
	
	\subsection{Dataset}
	
	In this paper, we use the most popular human pose estimation dataset, COCO. The COCO keypoint dataset \cite{lin2014microsoft} presents challenging images with multi-person pose of various body scales and occlusion patterns in unconstrained environments. It contains 200,000 images and 250,000 person samples. Each person instance is labelled with 17 joints. We train our models on \texttt{train2017}(includes 57K images and 150K person instances) with no extra data, and conduct ablation study on \texttt{val2017}. Then we test our models on 
	\texttt{test-dev}
	for comparison with the state-of-the-art methods. In evaluation, we use the metric of Object Keypoint Similarity (OKS) for COCO to report the model performance.
	
	\subsection{Implementation Details}
	
	For fair comparison, we follow the same training configuration as \cite{xiao2018simple} and \cite{sun2019deep} for ResNet and HRNet respectively. 
	To construct the localization subnet,
	we copy the conv5 stage, which spatial size is 1/32 to the input size, and the last three deconvolution layers for ResNet series networks, while copying the stage4, which has three high resolution modules, for HRNet series networks. For ablation study, we also add 128x96 input size in our experiment following \cite{zhang2019distribution}. And we set $N$ as 48, 192 and 432 corresponding to the three input sizes of 128x96, 256x192 and 384x288 during all of our experiment except the ablation study of $N$. During inference, we use person detectors of AP 56.4 and 60.9 for COCO \texttt{val2017} and \texttt{test-dev} respectively, while for pose estimation, we evaluate single model and only use flipping test strategy for testing argumentation. 
	
	\subsection{Ablation Studies} \label{sec:ablation}
	
	We use ResNet-50 backbone to perform ablation study on COCO 
	\texttt{val2017}.

	\setlength{\tabcolsep}{4pt}
	\begin{table} [t]
		\centering
		\caption{Ablation study on COCO 
			\texttt{val2017}
		}
		\label{table:ablation-self}
		\begin{tabular}{cccc}
			\hline\noalign{\smallskip}
			Method & Size & stage1 AP & stage2 AP \\
			\noalign{\smallskip}
			\hline
			\noalign{\smallskip}
			SBN & 128x96 & 59.3 & - \\
			Graph-PCNN & 128x96 & \textbf{61.1} & \textbf{64.6} \\
			SBN & 256x192 & 70.4 & - \\
			Graph-PCNN & 256x192 & \textbf{71.3} & \textbf{72.6} \\
			SBN & 384x288 & 72.2 & - \\
			Graph-PCNN & 384x288 & \textbf{72.7} & \textbf{73.6} \\
			\hline
		\end{tabular}
	\end{table}
	\setlength{\tabcolsep}{1.4pt}

	\pheadB{Two stage pose estimation framework.} Firstly, we evaluate the effectiveness of our proposed two stage pose estimation framework. As Table~\ref{table:ablation-self} shows, the stage2 of Graph-PCNN gives 5.3\%, 2.2\%, 1.4\% AP gain comparing to original simple baseline network(SBN) at the three input sizes, which demonstrates that our regression based two stage framework is more effective than decoding joint location from heatmap. Further more, we test the stage1 of Graph-PCNN which shares the same network architecture with SBN. It should be noted that training with Graph-PCNN can also boost the performance of heatmap, and 1.8\%, 0.9\%, 0.5\% AP gain are got as shown. That means we can also get considerable performance boosting without any extra computing cost during inference if we only use the stage1 of Graph-PCNN.

	\begin{figure} [t]
		\centering
		\includegraphics[width=1.0\textwidth]{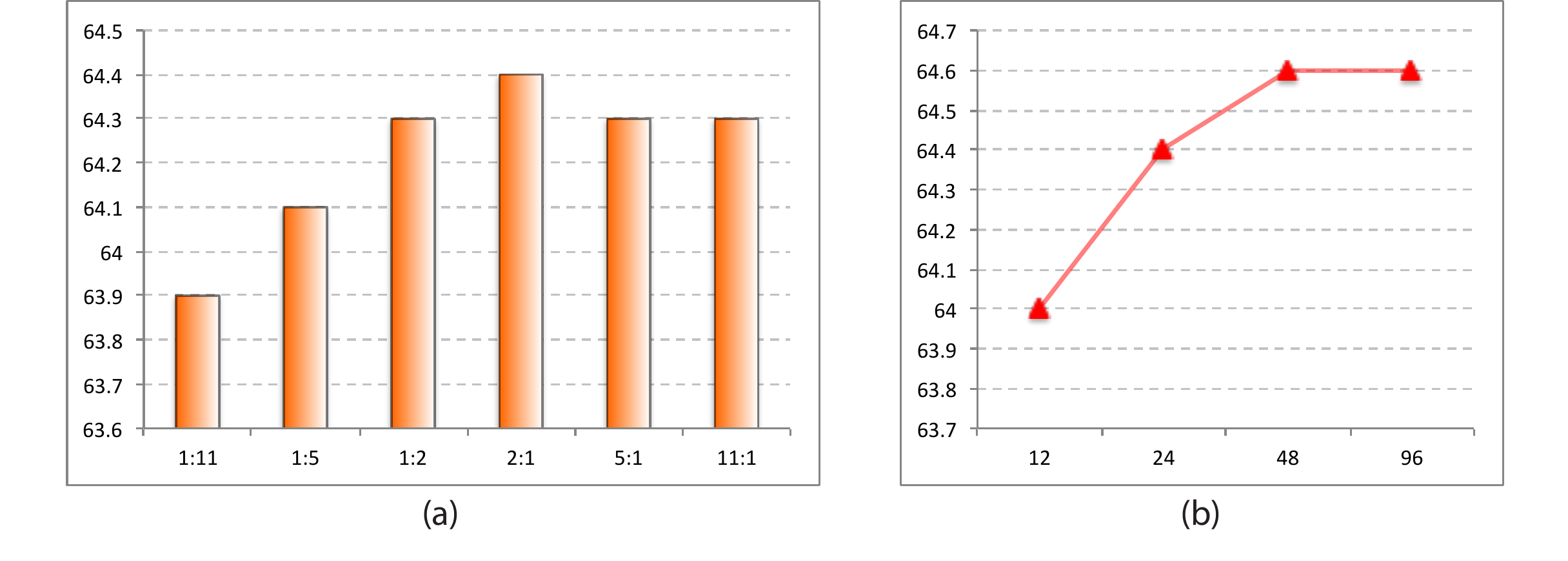}
		\caption{Influence of the proportion and total amount of guided points in sampling. (a) is the results on different proportions, while the x-axis represents the proportions between positive guided points and negative guided points. (b) is the results on different values of the total amount, while the x-axis represents the values of $N$.}
		\label{fig:ablation-sampling}
	\end{figure}

	\pheadB{Sampling strategy.} Secondly, we study the influence of the proportion of different kinds of guided points and the total amount of guided points $N$ based on ResNet-50 with 128x96 input size. In order to avoid exploring the proportion among all the three kinds of guided points, we simplify the proportion study by using only definite positive points and negative points, and then we set different proportion between them with $N$ unchanged. From the results shown in Fig.~\ref{fig:ablation-sampling} (a), we can come to that the proportion ranging from 1:2 to 2:1 is already appropriate, and the sampling strategy proposed in Section~\ref{sec:sampling} 
	can fit this proportion range in any 
	situation.
	In addition, we try different $N$ based on the strategy in Section~\ref{sec:sampling} and finally select 48 as the value of $N$ according to the results reported in Fig.~\ref{fig:ablation-sampling} (b).
	
	\pheadB{Graph pose refinement module.} Finally, we evaluate the contribution of the proposed graph pose refinement(GPR) module. In this study, 
	we compare proposed GPR with a struct-agnostic baseline module and several variants of GPR(GPR-va, GPR-vb, GPR-vc).
	GPR-va set $\omega_{k^{'}}=1$ for all $\{k^{'} | \vvec{s}_{k^{'}} \in \mathcal{N}(k)\}$ in (\ref{fml:computegcn}), GPR-vb set $\omega_{k^{'}}=\mathbbm{1}(R_{k^{'}})$ for $\{k^{'} | \vvec{s}_{k^{'}} \in \mathcal{N}(k), k^{'} \ne k\}$ with the heat response factor dropped, and GPR-vc dropped the guided points shuffling operation mentioned in Section~\ref{sec:detail}. The comparison results are displayed in Table~\ref{table:ablation-graph_coordinate_refinement}. We can see that GPR boosts the stage1 AP and stage2 AP by 0.4\% and 0.8\% respectively, comparing to the struct-agnostic baseline. 
	And the performance of GPR is better than all of its other variants, which reveals the importance of parameter $\omega_{k^{'}}$ and the guided points shuffling operation. Especially, the reliability factor $\mathbbm{1}(R_{k^{'}})$ affects the performance greatly.
	Thus, we believe that GPR can refine the feature of a guided point by taking advantage of the supervision signal of its neighbour keypoint which is good located, as we supposed in Section~\ref{sec:gcn}.
	
	\setlength{\tabcolsep}{4pt}
	\begin{table} [t]
		\centering
		\caption{Effectiveness of the graph pose refinement(GPR) module.}
		\label{table:ablation-graph_coordinate_refinement}
		\begin{tabular}{cccc}
			\hline\noalign{\smallskip}
			Method & Size & stage1 AP & stage2 AP \\
			\noalign{\smallskip}
			\hline
			\noalign{\smallskip}
			struct-agnostic & 128x96 & 60.7 & 63.8 \\
			GPR-va & 128x96 & 61.2 & 52.1 \\
			GPR-vb & 128x96 & 61.1 & 64.5 \\
			GPR-vc & 128x96 & 60.8 & 64.3 \\
			GPR & 128x96 & 61.1 & \textbf{64.6} \\
			\hline
		\end{tabular}
	\end{table}
	\setlength{\tabcolsep}{1.4pt}
	
	\subsection{Comparison with Other Methods with Coordinate Refinement}
	
	DARK\cite{zhang2019distribution} is a state-of-the-art method which improved traditional decoding by a more precise refinement based on Taylor-expansion. We follow the training settings of DARK and compare our refinement results with it. From Table~\ref{table:ablation-dark} we can observe that our 
	Graph-PCNN
	generally outperforms DARK over different network architecture and input size. This suggests that regression based refinement predicts coordinate more precise than analyzing the distribution of response signal from heatmap, as the response signal itself may not satisfy gaussian distribution strictly because of complex human pose and image context while regression is regardless of these drawback.
	
	\setlength{\tabcolsep}{4pt}
	\begin{table} [t]
		\centering
		\caption{Comparison with distribution-aware coordinate representation of keypoint(DARK) on COCO \texttt{val2017}.}
		\label{table:ablation-dark}
		\begin{tabular}{ccccccccc}
			\hline\noalign{\smallskip}
			Method & Backbone & Size & \emph{AP} & \emph{$AP^{50}$} & \emph{$AP^{75}$} & \emph{$AP^M$} & \emph{$AP^L$} & \emph{AR} \\
			\noalign{\smallskip}
			\hline
			\noalign{\smallskip}
			DARK & R50 & 128x96 & 62.6 & 86.1 & 70.4 & 60.4 & 67.9 & 69.5 \\
			Graph-PCNN & R50 & 128x96 & \textbf{64.6} & \textbf{86.4} & \textbf{72.7} & \textbf{62.4} & \textbf{70.1} & \textbf{71.5} \\
			\hline
			\noalign{\smallskip}
			DARK & R101 & 128x96 & 63.2 & 86.2 & 71.1 & 61.2 & 68.5 & 70.0 \\
			Graph-PCNN & R101 & 128x96 & \textbf{64.8} & \textbf{86.6} & \textbf{73.1} & \textbf{62.6} & \textbf{70.3} & \textbf{71.7} \\
			\hline
			\noalign{\smallskip}
			DARK & R152 & 128x96 & 63.1 & 86.2 & 71.6 & 61.3 & 68.1 & 70.0 \\
			Graph-PCNN & R152 & 128x96 & \textbf{66.1} & \textbf{87.2} & \textbf{74.6} & \textbf{64.1} & \textbf{71.5} & \textbf{73.0} \\
			\hline
			\noalign{\smallskip}
			DARK & HR32 & 128x96 & 70.7 & 88.9 & 78.4 & 67.9 & 76.6 & 76.7 \\
			Graph-PCNN & HR32 & 128x96 & \textbf{71.5} & \textbf{89.0} & \textbf{79.0} & \textbf{68.4} & \textbf{77.6} & \textbf{77.3} \\
			\hline
			\noalign{\smallskip}
			DARK & HR48 & 128x96 & 71.9 & 89.1 & 79.6 & 69.2 & 78.0 & 77.9 \\
			Graph-PCNN & HR48 & 128x96 & \textbf{72.8} & \textbf{89.2} & \textbf{80.1} & \textbf{69.9} & \textbf{79.0} & \textbf{78.6} \\
			\hline
			\noalign{\smallskip}
			\hline
			\noalign{\smallskip}
			DARK & HR32 & 256x192 & 75.6 & 90.5 & 82.1 & 71.8 & 82.8 & 80.8 \\
			Graph-PCNN & HR32 & 256x192 & \textbf{76.2} & \textbf{90.3} & \textbf{82.6} & \textbf{72.5} & \textbf{83.2} & \textbf{81.2} \\
			\hline
			\noalign{\smallskip}
			DARK & HR32 & 384x288 & 76.6 & 90.7 & 82.8 & 72.7 & 83.9 & 81.5 \\
			Graph-PCNN & HR32 & 384x288 & \textbf{77.2} & \textbf{90.7} & \textbf{83.6} & \textbf{73.5} & \textbf{84.0} & \textbf{82.1} \\
			\hline
		\end{tabular}
	\end{table}
	\setlength{\tabcolsep}{1.4pt}

	PoseFix\cite{moon2019posefix} is a model-agnostic method which refines a existing pose result from any other method by a independent model. A coarse-to-fine coordinate estimation schedule ended by coordinate calculation following integral loss\cite{sun2018integral} is used to enhance the precision. We conduct comparison with PoseFix by using same backbone and input size with its model from refinement stage and the performance of human detectors for these two methods are comparable, AP 55.3 vs 56.4 for PoseFix(using CPN) and our 
	Graph-PCNN
	respectively. As illustraed in Table~\ref{table:ablation-posefix}, we achieve a competable result with PoseFix, but PoseFix included input from CPN which need an extra R50 network while our method only need an extra R50 conv5 stage as refinement branch.

	\setlength{\tabcolsep}{4pt}
	\begin{table} [t]
		\centering
		\caption{Comparison with model-agnostic human pose refinement network(PoseFix) on COCO \texttt{val2017}.}
		\label{table:ablation-posefix}
		\begin{tabular}{ccccccccc}
			\hline\noalign{\smallskip}
			Method & Backbone & Size & \emph{AP} & \emph{$AP^{50}$} & \emph{$AP^{75}$} & \emph{$AP^M$} & \emph{$AP^L$} & \emph{AR} \\
			\noalign{\smallskip}
			\hline
			\noalign{\smallskip}
			PoseFix & R50 & 256x192 & 72.1 & 88.5 & 78.3 & 68.6 & 78.2 & - \\
			Graph-PCNN & R50 & 256x192 & \textbf{72.6} & \textbf{89.1} & \textbf{79.3} & \textbf{69.1} & \textbf{79.7} & 78.1 \\
			\hline
		\end{tabular}
	\end{table}
	\setlength{\tabcolsep}{1.4pt}

	\subsection{Comparison to State of the Art}
	We compare our Graph-PCNN with other top-performed methods on COCO \texttt{test-dev}. As Table~\ref{table:sota-test-dev} reports, our method with HR48 backbone at the input size of 384x288 achieves the best AP(76.8), and improves HR48 with the same input size(75.5) by a large margin(+1.3). Mean while, It also outperforms other competitors with same backbone and input size settings, such as DARK(76.2), UDP(76.5) and PoseFix(76.7), which illustrates the advantages of our method.

	\setlength{\tabcolsep}{4pt}
	\begin{table} [t]
		\centering
		\caption{Comparison with the state-of-the-arts methods on COCO \texttt{test-dev}.}
		\label{table:sota-test-dev}
		\resizebox{1.0\textwidth}{!}{
			\begin{tabular}{ccccccccc}
				\hline\noalign{\smallskip}
				Method & Backbone & Size & \emph{AP} & \emph{$AP^{50}$} & \emph{$AP^{75}$} & \emph{$AP^M$} & \emph{$AP^L$} & \emph{AR} \\
				\noalign{\smallskip}
				\hline
				\noalign{\smallskip}
				CMU-Pose\cite{cao2017realtime} & - & - & 61.8 & 84.9 & 67.5 & 57.1 & 68.2 & 66.5 \\
				Mask-RCNN\cite{he2017mask} & R50-FPN & - & 63.1 & 87.3 & 68.7 & 57.8 & 71.4 & - \\
				G-RMI\cite{papandreou2017towards} & R101 & 353x257 & 64.9 & 85.5 & 71.3 & 62.3 & 70.0 & 69.7 \\
				AE\cite{newell2017associative} & - & 512x512 & 65.5 & 86.8 & 72.3 & 60.6 & 72.6 & 70.2 \\
				Integral Pose\cite{sun2018integral} & R101 & 256x256 & 67.8 & 88.2 & 74.8 & 63.9 & 74.0 & - \\
				CPN\cite{chen2018cascaded} & ResNet-Inception & 384x288 & 72.1 & 91.4 & 80.0 & 68.7 & 77.2 & 78.5 \\
				RMPE\cite{fang2017rmpe} & PyraNet\cite{yang2017pyramid} & 320x256 & 72.3 & 89.2 & 79.1 & 68.0 & 78.6 & - \\
				CFN\cite{huang2017coarse} & - & - & 72.6 & 86.1 & 69.7 & 78.3 & 64.1 & - \\
				CPN(ensemble)\cite{chen2018cascaded} & ResNet-Inception & 384x288 & 73.0 & 91.7 & 80.9 & 69.5 & 78.1 & 79.0 \\
				Posefix\cite{moon2019posefix} & R152+R152 & 384x288 & 73.6 & 90.8 & 81.0 & 70.3 & 79.8 & 79.0 \\
				CSM+SCARB\cite{su2019multi} & R152 & 384x288 & 74.3 & 91.8 & 81.9 & 70.7 & 80.2 & 80.5 \\
				CSANet\cite{yu2019context} & R152 & 384x288 & 74.5 & 91.7 & 82.1 & 71.2 & 80.2 & 80.7 \\
				MSPN\cite{li2019rethinking} & MSPN & 384x288 & 76.1 & 93.4 & 83.8 & 72.3 & 81.5 & 81.6 \\
				\hline
				\noalign{\smallskip}
				\hline
				\noalign{\smallskip}
				Simple Base\cite{xiao2018simple} & R152 & 384x288 & 73.7 & 91.9 & 81.1 & 70.3 & 80.0 & 79.0 \\
				UDP\cite{huang2019devil} & R152 & 384x288 & 74.7 & 91.8 & 82.1 & 71.5 & 80.8 & 80.0 \\
				\textbf{Graph-PCNN} & R152 & 384x288 & \textbf{75.1} & \textbf{91.8} & \textbf{82.3} & \textbf{71.6} & \textbf{81.4} & \textbf{80.2} \\
				\hline
				\noalign{\smallskip}
				HRNet\cite{sun2019deep} & HR32 & 384x288 & 74.9 & 92.5 & 82.8 & 71.3 & 80.9 & 80.1 \\
				UDP\cite{huang2019devil} & HR32 & 384x288 & 76.1 & 92.5 & 83.5 & 72.8 & 82.0 & 81.3 \\
				\textbf{Graph-PCNN} & HR32 & 384x288 & \textbf{76.4} & \textbf{92.5} & \textbf{83.8} & \textbf{72.9} & \textbf{82.4} & \textbf{81.3} \\
				\hline
				\noalign{\smallskip}
				HRNet\cite{sun2019deep} & HR48 & 384x288 & 75.5 & 92.5 & 83.3 & 71.9 & 81.5 & 80.5 \\
				DARK\cite{zhang2019distribution} & HR48 & 384x288 & 76.2 & 92.5 & 83.6 & 72.5 & 82.4 & 81.1 \\
				UDP\cite{huang2019devil} & HR48 & 384x288 & 76.5 & \textbf{92.7} & 84.0 & 73.0 & 82.4 & 81.6 \\
				PoseFix\cite{moon2019posefix} & HR48+R152 & 384x288 & 76.7 & 92.6 & 84.1 & 73.1 & 82.6 & 81.5 \\
				\textbf{Graph-PCNN} & HR48 & 384x288 & \textbf{76.8} & 92.6 & \textbf{84.3} & \textbf{73.3} & \textbf{82.7} & \textbf{81.6} \\
				\hline
			\end{tabular}
		}
	\end{table}
	\setlength{\tabcolsep}{1.4pt}

	\section{Conclusions}
	
	In this paper, we propose a two stage human pose estimator for the top-down pose estimation network, which improves the overall 
	localization
	performance by introducing different features for rough and accurate localization. Meanwhile, a graph pose refinement module is proposed to refine the feature for pose regression by taking the relationship between keypoints into account, which boosts the performance of our two stage pose estimator further. Our proposed method is 
	model-agnostic
	and can be added on most of the mainstream backbone. Even better, more improvement can be explored by drawing on the successful experience of the two stage detection framework in the future.
	
	%
	%
	\bibliographystyle{splncs04}
	\bibliography{egbib}

\begin{thebibliography}{10}
\providecommand{\url}[1]{\texttt{#1}}
\providecommand{\urlprefix}{URL }
\providecommand{\doi}[1]{https://doi.org/#1}

\bibitem{andriluka20142d}
Andriluka, M., Pishchulin, L., Gehler, P., Schiele, B.: 2d human pose
  estimation: New benchmark and state of the art analysis. In: CVPR (2014)

\bibitem{belagiannis2015robust}
Belagiannis, V., Rupprecht, C., Carneiro, G., Navab, N.: Robust optimization
  for deep regression. In: Proceedings of the IEEE international conference on
  computer vision. pp. 2830--2838 (2015)

\bibitem{belagiannis2016recurrent}
Belagiannis, V., Zisserman, A.: Recurrent human pose estimation. In: FG (2017)

\bibitem{bulat2016human}
Bulat, A., Tzimiropoulos, G.: Human pose estimation via convolutional part
  heatmap regression. In: European Conference on Computer Vision. pp. 717--732.
  Springer (2016)

\bibitem{cao2017realtime}
Cao, Z., Simon, T., Wei, S.E., Sheikh, Y.: Realtime multi-person 2d pose
  estimation using part affinity fields. In: CVPR (2017)

\bibitem{carreira2015human}
Carreira, J., Agrawal, P., Fragkiadaki, K., Malik, J.: Human pose estimation
  with iterative error feedback. In: CVPR (2016)

\bibitem{chen2014articulated}
Chen, X., Yuille, A.L.: Articulated pose estimation by a graphical model with
  image dependent pairwise relations. In: NeurIPS (2014)

\bibitem{chen2018cascaded}
Chen, Y., Wang, Z., Peng, Y., Zhang, Z., Yu, G., Sun, J.: Cascaded pyramid
  network for multi-person pose estimation. In: CVPR (2018)

\bibitem{chen2017adversarial}
Chen, Y., Shen, C., Wei, X.S., Liu, L., Yang, J.: Adversarial posenet: A
  structure-aware convolutional network for human pose estimation. In: ICCV
  (2017)

\bibitem{cho2013adaptive}
Cho, N.G., Yuille, A.L., Lee, S.W.: Adaptive occlusion state estimation for
  human pose tracking under self-occlusions. Pattern Recognition
  \textbf{46}(3),  649--661 (2013)

\bibitem{chu2016structured}
Chu, X., Ouyang, W., Li, H., Wang, X.: Structured feature learning for pose
  estimation. In: CVPR (2016)

\bibitem{chu2017multi}
Chu, X., Yang, W., Ouyang, W., Ma, C., Yuille, A.L., Wang, X.: Multi-context
  attention for human pose estimation. In: CVPR (2017)

\bibitem{fan2015combining}
Fan, X., Zheng, K., Lin, Y., Wang, S.: Combining local appearance and holistic
  view: Dual-source deep neural networks for human pose estimation. In: CVPR
  (2015)

\bibitem{fang2017rmpe}
Fang, H.S., Xie, S., Tai, Y.W., Lu, C.: Rmpe: Regional multi-person pose
  estimation. In: ICCV (2017)

\bibitem{fieraru2018learning}
Fieraru, M., Khoreva, A., Pishchulin, L., Schiele, B.: Learning to refine human
  pose estimation. In: CVPR (2018)

\bibitem{gkioxari2016chained}
Gkioxari, G., Toshev, A., Jaitly, N.: Chained predictions using convolutional
  neural networks. In: ECCV (2016)

\bibitem{he2017mask}
He, K., Gkioxari, G., Doll{\'a}r, P., Girshick, R.: Mask r-cnn. In: ICCV (2017)

\bibitem{huang2019devil}
Huang, J., Zhu, Z., Guo, F., Huang, G.: The devil is in the details: Delving
  into unbiased data processing for human pose estimation. arXiv preprint
  arXiv:1911.07524  (2019)

\bibitem{huang2017coarse}
Huang, S., Gong, M., Tao, D.: A coarse-fine network for keypoint localization.
  In: ICCV (2017)

\bibitem{ke2018multi}
Ke, L., Chang, M.C., Qi, H., Lyu, S.: Multi-scale for human pose estimation.
  In: ECCV (2018)

\bibitem{li2019rethinking}
Li, W., Wang, Z., Yin, B., Peng, Q., Du, Y., Xiao, T., Yu, G., Lu, H., Wei, Y.,
  Sun, J.: Rethinking on multi-stage networks for human pose estimation. arXiv
  preprint arXiv:1901.00148  (2019)

\bibitem{liang2014expressive}
Liang, Z., Wang, X., Huang, R., Lin, L.: An expressive deep model for human
  action parsing from a single image. In: ICME. IEEE (2014)

\bibitem{lifshitz2016human}
Lifshitz, I., Fetaya, E., Ullman, S.: Human pose estimation using deep
  consensus voting. In: ECCV (2016)

\bibitem{lin2014microsoft}
Lin, T.Y., Maire, M., Belongie, S., Hays, J., Perona, P., Ramanan, D.,
  Doll{\'a}r, P., Zitnick, C.L.: Microsoft coco: Common objects in context. In:
  ECCV (2014)

\bibitem{liu2018cascaded}
Liu, W., Chen, J., Li, C., Qian, C., Chu, X., Hu, X.: A cascaded inception of
  inception network with attention modulated feature fusion for human pose
  estimation. In: AAAI (2018)

\bibitem{moon2019posefix}
Moon, G., Chang, J.Y., Lee, K.M.: Posefix: Model-agnostic general human pose
  refinement network. In: CVPR (2019)

\bibitem{newell2017associative}
Newell, A., Huang, Z., Deng, J.: Associative embedding: End-to-end learning for
  joint detection and grouping. In: NeurIPS (2017)

\bibitem{newell2016stacked}
Newell, A., Yang, K., Deng, J.: Stacked hourglass networks for human pose
  estimation. In: ECCV (2016)

\bibitem{ning2017tmm}
Ning, G., Zhang, Z., He, Z.: Knowledge-guided deep fractal neural networks for
  human pose estimation. IEEE Transactions on Multimedia  \textbf{PP}(99),
  ~1--1 (2017)

\bibitem{papandreou2017towards}
Papandreou, G., Zhu, T., Kanazawa, N., Toshev, A., Tompson, J., Bregler, C.,
  Murphy, K.: Towards accurate multi-person pose estimation in the wild. In:
  CVPR (2017)

\bibitem{pishchulin2013poselet}
Pishchulin, L., Andriluka, M., Gehler, P., Schiele, B.: Poselet conditioned
  pictorial structures. In: CVPR (2013)

\bibitem{su2019multi}
Su, K., Yu, D., Xu, Z., Geng, X., Wang, C.: Multi-person pose estimation with
  enhanced channel-wise and spatial information. In: CVPR (2019)

\bibitem{sun2019deep}
Sun, K., Xiao, B., Liu, D., Wang, J.: Deep high-resolution representation
  learning for human pose estimation. In: CVPR (2019)

\bibitem{sun2017compositional}
Sun, X., Shang, J., Liang, S., Wei, Y.: Compositional human pose regression.
  In: ICCV (2017)

\bibitem{sun2018integral}
Sun, X., Xiao, B., Wei, F., Liang, S., Wei, Y.: Integral human pose regression.
  In: ECCV (September 2018)

\bibitem{tang2018deeply}
Tang, W., Yu, P., Wu, Y.: Deeply learned compositional models for human pose
  estimation. In: ECCV (2018)

\bibitem{tompson2014joint}
Tompson, J.J., Jain, A., LeCun, Y., Bregler, C.: Joint training of a
  convolutional network and a graphical model for human pose estimation. In:
  NeurIPS (2014)

\bibitem{toshev2014deeppose}
Toshev, A., Szegedy, C.: Deeppose: Human pose estimation via deep neural
  networks. In: CVPR (2014)

\bibitem{wang2013approach}
Wang, C., Wang, Y., Yuille, A.L.: An approach to pose-based action recognition.
  In: CVPR (2013)

\bibitem{wei2016convolutional}
Wei, S.E., Ramakrishna, V., Kanade, T., Sheikh, Y.: Convolutional pose
  machines. In: CVPR (2016)

\bibitem{chu2016crf}
Xiao, Ouyang, W., Wang, X., et~al.: Crf-cnn: Modeling structured information in
  human pose estimation. In: NeurIPS (2016)

\bibitem{xiao2018simple}
Xiao, B., Wu, H., Wei, Y.: Simple baselines for human pose estimation and
  tracking. In: ECCV (2018)

\bibitem{xiaohan2015joint}
Xiaohan~Nie, B., Xiong, C., Zhu, S.C.: Joint action recognition and pose
  estimation from video. In: CVPR (2015)

\bibitem{yan2018spatial}
Yan, S., Xiong, Y., Lin, D.: Spatial temporal graph convolutional networks for
  skeleton-based action recognition. In: AAAI (2018)

\bibitem{yang2017pyramid}
Yang, W., Li, S., Ouyang, W., Li, H., Wang, X.: Learning feature pyramids for
  human pose estimation. In: ICCV (2017)

\bibitem{yang2016end}
Yang, W., Ouyang, W., Li, H., Wang, X.: End-to-end learning of deformable
  mixture of parts and deep convolutional neural networks for human pose
  estimation. In: CVPR (2016)

\bibitem{yang2012articulated}
Yang, Y., Ramanan, D.: Articulated human detection with flexible mixtures of
  parts. IEEE transactions on pattern analysis and machine intelligence
  \textbf{35}(12),  2878--2890 (2012)

\bibitem{yu2019context}
Yu, D., Su, K., Geng, X., Wang, C.: A context-and-spatial aware network for
  multi-person pose estimation. arXiv preprint arXiv:1905.05355  (2019)

\bibitem{zhang2019distribution}
Zhang, F., Zhu, X., Dai, H., Ye, M., Zhu, C.: Distribution-aware coordinate
  representation for human pose estimation. arXiv preprint arXiv:1910.06278
  (2019)

\bibitem{zhang2019human}
Zhang, H., Ouyang, H., Liu, S., Qi, X., Shen, X., Yang, R., Jia, J.: Human pose
  estimation with spatial contextual information. arXiv preprint
  arXiv:1901.01760  (2019)

\end{thebibliography}
\end{document}